# HashSet - A Dataset For Hashtag Segmentation

**Prashant Kodali[†], Akshala Bhatnagar[*], Naman Ahuja[†]**
**Manish Shrivastava[†], Ponnurangam Kumaraguru[†]**
[†]IIIT Hyderabad
prashant.kodali@research.iiit.ac.in, naman.ahuja@students.iiit.ac.in,
{m.shrivastava, pk.guru}@iiit.ac.in
[*]IIIT Delhi
akshala18012@iiitd.ac.in

**Abstract**
Hashtag segmentation is the task of breaking a hashtag into its constituent tokens. Hashtags often encode the essence of user-generated posts, along with information like topic and sentiment, which are useful in downstream tasks. Hashtags prioritize brevity and are written in unique ways - transliterating and mixing languages, spelling variations, creative named entities. Benchmark datasets used for the hashtag segmentation task - STAN, BOUN - are small in size and extracted from a single set of tweets. However, datasets should reflect the variations in writing styles of hashtags and also account for domain and language specificity, failing which the results will misrepresent model performance. We argue that model performance should be assessed on a wider variety of hashtags, and datasets should be carefully curated. To this end, we propose HashSet, a dataset comprising of: a) 1.9k manually annotated dataset; b) 3.3M loosely supervised dataset. HashSet dataset is sampled from a different set of tweets when compared to existing datasets and provides an alternate distribution of hashtags to build and validate hashtag segmentation models. We show that the performance of SOTA models for Hashtag Segmentation drops substantially on proposed dataset, indicating that the proposed dataset provides an alternate set of hashtags to train and assess models. Datasets and results are released publicly and can be accessed from `https://github.com/prashantkodali/HashSet`

## 1. Introduction

Hashtags have become ubiquitous across user-generated content on the Internet. Hashtags often encapsulate the gist, emotion, and sentiment cues of the text (Qadir and Riloff, 2014), and have been demonstrated to be useful in downstream tasks like text classification (Belainine et al., 2016). Hashtags, however, pose a challenge in automatic processing because of their unsegmented nature. To leverage hashtags in downstream task, hashtags have to be broken into their constituent tokens, a task which is called Hashtag Segmentation or Hashtag Decomposition.

Segmenting a hashtag is akin to a word segmentation problem. Hashtags show specific quirks like spelling variations (e.g., #letzgooo), romanization of native language words (e.g., #sabkasaath), camel case (e.g., #WeLoveApples) and presence of special characters (e.g., #We_love_apples@1). Presence of such quirks in hashtags makes the solution of hashtag segmentation non-trivial and slightly different from word segmentation e.g., "letsgo" can be segmented easily compared to "letzgoo" since the later has non-cannonical spellings.

Recently proposed methods (Maddela et al., 2019; Rodrigues et al., 2021) have leveraged the power of language models and combined them with neual ranking models to segment hashtags. STAN (Bansal et al., 2015), BOUN (Çelebi and Özgür, 2016) are the popular benchmark datasets for Hashtag segmentation. Test set sizes for STAN and BOUN datasets are 1012, and 999 respectively. Small-sized datasets make it harder to train supervised models and aren't representative of the large variety of hashtags that are observed across user-generated content. Model's performance reported on such datasets could be misleading and could drop down if tested on hashtags from a different geographical location and different domain. It is, thus, pertinent to construct datasets consisting of non-trivial samples and samples which are often misclassified by SOTA model. Hashtags often are written in camelcase, e.g., #WeLoveApples; or use underscores to seperate tokens, e.g., #We_love_apples. Hashtags written using such commonly occurring patterns can be segmented using simple hand-crafted methods instead of relying on the power of large Language models and complex machine learning models. Hashtags which could be segmented using such strategies are relatively easy for the model to segment since they exhibit peculiar and frequent patterns.

We propose that benchmark hashtag segmentation datasets should prioritize non-trivial cases such that model performance is truly representative of task performance. To effectively evaluate hashtags segmentation models, the benchmark datasets should reflect the variety in hashtags in terms of language variety and named entities.

As a primary contribution of our work, we propose **HashSet**, a new dataset for hashtag segmentation. HashSet dataset consists of two components:

- **HashSet-Manual** - 1,901 hashtags manually annotated for consituent segments, named entities, and whether or not hashtag contains non-english tokens.

- **HashSet-Distant** - 332,166 hashtags segmented automatically using the camel case cues.

To the best of our knowledge, HashSet-Manual is the only publicly available hashtag segmentation dataset, which has named entity annotation, along with binary annotation for the presence/absence of non-English tokens. HashSet-Distant is a large collection of camel-case hashtags that are segmented automatically leveraging the case information, forming the largest distant supervision dataset for hashtag segmentation.

We also report the performance of models proposed by Maddela et al. (2019), Rodrigues et al. (2021), which are SOTA models to the best of our knowledge. We report results on HashSet along with STAN and BOUN and compare their performance.

The remainder of this paper is organized as follows. In Section 2, we discuss the background work and language resources. In Section 3, we introduce our dataset and contrast it with the existing datasets. In Section 4, we present SOTA models on the datasets and compare them across datasets. Finally, in Section 5, we present our conclusions, list limitations, and propose avenues for future work.

## 2. Related Work

For a majority of hashtag datasets, source of hashtags is Stanford Sentiment Analysis Dataset (Go et al., 2009). (Bansal et al., 2015) extracted hashtags from the Stanford Sentiment Analysis Dataset and manually annotated hashtags for their segments, and was further extended by (Çelebi and Özgür, 2016). For the rest of the paper, we refer to them as $STAN_{small}$ and $STAN_{dev}$, respectively. Çelebi and Özgür (2016) created BOUN dataset by manually segmenting hashtags obtained by randomly querying Twitter API for movies, tv-shows, titles, people names, etc. Maddela et al. (2019) proposed $STAN_{large}$, comprising of 12,594 hashtags manually annotated for their segments using crowd sourcing. Maddela et al. (2019) note that nearly 33% of hashtags have named entities, and 47.1% single-token and 52.9% were multi-word hashtags in $STAN_{large}$ dataset. However, the annotations for named entities aren't publicly for available $STAN_{large}$.

Proposed models relied on lexical resources and/or language models to generate candidate segmentations and rank the candidates. Maddela et al. (2019) proposed a model which used statistical Language Models to generate candidates, which are further re-ranked using neural architectures. Statistical language models were trained on a large collection of English tweets. Proposed model achieved SOTA performance on $STAN_{small}$ and $STAN_{large}$ datasets. Rodrigues et al. (2021) proposed a zero-shot architecture, Hashformer, which leverages ensemble of transformer-based language models, and re-ranking to generate segments. Both models relied on language models to generate candidates. Hashformers reported SOTA performance on $STAN_{small}$ and BOUN dataset. Hashformer is a zero-shot method but uses annotated data to tune hyperparameters of the model. Language models have domain and language specificity. Efficacy of the hashtag segmentation algorithm will change depending on the geographical location and language used in the user-generated post. To overcome the limitations of prior datasets, we propose HashSet. In the following section, we introduce our dataset, HashSet, and compare it against the existing datasets.

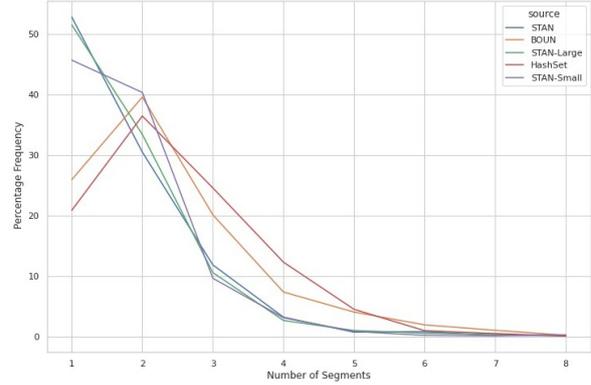

Figure 1: Distribution of segments across datasets. HashSet has higher proportion of multiple segment hashtags as compared to STAN and BOUN

## 3. HashSet - Dataset Description

We construct HashSet dataset from a collection of tweets. We annotate a subset of these hashtags to create Hashtag-Manual. We segregate the camel cased hashtags, use regular expression rules to create Hashset-Distant. Data collection, annotation methodology, and dataset statistics are detailed in the following subsections.

### 3.1. Data Collection

To create a large set of hashtags, we used Twitter API in the following ways: a) queried Twitter API for trending hashtags across different locations for the period May-October 2021; b) hashtags from a collection of tweets for trending hashtags during the period of April - May 2019. Two sets of collections help us account for numerous non-trending hashtags. We collected 841,520 unique hashtags in total. We filter out hashtags that aren't written in roman script, ending up with 731,357 hashtags, out of which 319,497 hashtags were in camel case.

### 3.2. Annotation Process

We used LabelStudio [1], an annotation tool that is used to create data resources for text, images, audio, video. We randomly sample hashtags from the aforementioned collection of hashtags, and three annotators annotated a total of 1,901 hashtags from the collected set.

---
[1] https://labelstud.io/

|  | **Datasets** | | | | | |
| Parameter | STAN-Dev | STAN-Small | STAN-Large | BOUN | HashSet-Manual | HashSet-Distant |
| --- | --- | --- | --- | --- | --- | --- |
| Number of Hashtags | 1012 | 1108 | 11965 | 999 | 1901 | 332166 |
| Avg. Hashtag Length | 8.49 | 8.9 | 8.58 | 11.3 | 12.68 | 14.69 |
| Avg. Number of Segments | 1.75 | 1.78 | 1.74 | 2.37 | 2.49 | 2.8 |
| Num of Single Token Hashtags | 532 (53%) | 453 (41%) | 4749 (40%) | 258 (26%) | 396 (20.8%) | 0 |
| Num of Camel Case Hashtags | 134 (13.3%) | 1441 (12.0%) | 108 (9.8%) | 278 (27.8%) | 0 | 332166(100%) |
| Num of Non-English Tokens | - | - | - | - | 236 (12.4%) | - |
| Num of Named Entities | - | - | - | - | 1414 (74.4%) | - |

Table 1: Relevant Statistics from Datasets used in this study. We compare statistics (avg. hashtag length, avg. number of segments, single token hashtags) across these datasets to argue that HashSet serves a better corpus to gauge the efficacy of hashtag-segmentation models.

For every hashtag, the top 10 best segmentations, generated using the baseline model as proposed by Maddela et al. (2019), were presented to annotators. The annotator either chooses from the given 10 options and if the correct segmentation isn't present among the options, then the annotator writes their own segmentation. We used the predictions of the baseline model to speed up the annotation process. There are some hashtags that the annotators are unable to segment with certainty and mark as ambiguous; 89 such hashtags were found in the annotation process and were excluded.

In our preliminary analysis of the results from the baseline models, we inferred that the hashtags with named entities were segmented incorrectly. We also wanted to gauge the baseline method's capability of segmenting hashtags that had non-English tokens in them. To this end, apart from the correct segmentation, we labeled all the named entities for every hashtag. In addition to correct segmentation annotation, for every hashtag, we recorded annotator's responses to the following three questions :

- Does the hashtag contain a named entity?

- Does the hashtag contain non-English tokens?

Among the annotated hashtags for which the correct segmentation wasn't in the top 10 best ones (447 out of 1901), 354 of them contained named entities with an average of 1.23 named entities per hashtag supporting our initial hypothesis that presence of named entities lead to incorrect segmentation. In the collected hashtags, we find, out of the annotated 1901, 1414 contain named entities, 236 non-English tokens, as shown in Table 1. Some hashtags also contain more than one named entity (e.g., #Bjp4Bihar). On average, HashSet-Manual contains 1.10 (min = 0, max = 4) named entities per hashtag and an average of 2.42 (min = 1, max = 10) segments per hashtag. We argue that since HashSet has higher degree of named entities, it is comparatively tougher and a robust benchmark for Hashtag segmentation. HashSet contains relatively fewer single token hashtags. HashSet also has a higher mean hashtag length and segments as compared to STAN and BOUN, which points to the discernment of our dataset.

### 3.3. HashSet- Distant

Manual annotation of the hashtags is a time-consuming process. In order to create a large corpus of segmented hashtags, we leverage camel cased hashtags to create loosely supervised data at scale, which can be used to train and test supervised models. For the collected hashtags, we identify the total number of hashtags that are written in camel case and/or use underscores. Nearly 43% of the collected hashtags are written in camel case, and 3% of hashtags have underscores. We take the camel cased hashtags and construct the HashSet-Distant dataset.

For hashtags in HashSet-Distant, we use the camel case points to split the hashtags and create the loosely supervised hashtag segmentation data. On manual analysis, we infer that there are camel cased hashtags that can be segmented correctly just by splitting at camel cases. We implement regular expression-based patterns to split the camel case hashtags. In addition to the camel cased hashtags, we also store lower cased version of hashtags and their segments to estimate if lower casing makes it harder for SOTA model to perform (discussed in Section 4).

There are certain cases where such a regular expression-based method would fail; e.g., #CostofViraatvacation would be segmented as Costof Viraatvacation whereas the correct segmentation would be Cost of Viraat vacation. However, on our manual analysis of the resulting segments, we notice that they are minuscule. Further, we argue that if a hashtag is segmented using camel case cues, it would still help in increasing the performance of the underlying model.

### 4. Results & Error Analysis

To demonstrate the quality of the dataset, we report the performance of two recent SOTA models: a) Multi-task Pairwise Neural Ranking (MPNR) proposed by (Maddela et al., 2019); b) Hashformer proposed by (Rodrigues et al., 2021). We compare the performance of models on HashSet, along with other benchmark datasets - BOUN, $STAN_{dev}$, $STAN_{small}$, $STAN_{large}$. We use the publicly available implemen-tation of the SOTA models[2,3] and reproduce the results on all datasets for further analysis. For the MPNR model, we reproduce the results using the language models released by authors, and for Hashformers we use the publicly available versions of GPT-2, BERT.

| Architecture | Dataset | Accuracy @ top -n | | | | | |
|---|---|---|---|---|---|---|---|
| | | n=1 | n=2 | n=5 | n=7 | n=9 | n=10 |
| MPNR | BOUN | 81.6 | 88.09 | 90.29 | 90.69 | 90.69 | 90.69 |
| | STAN-Dev | 73.12 | 78.16 | 81.92 | 82.71 | 82.81 | 82.81 |
| | STAN-Small | 82.76 | 86.19 | 86.82 | 86.82 | 86.82 | 86.82 |
| | STAN-Large | 63.78 | 73.10 | 74.73 | 74.75 | 74.75 | 74.75 |
| | HashSet-Manual | 41.93 | 45.98 | 47.5 | 47.71 | 47.71 | 47.71 |
| Hashformers | BOUN | 83.68 | 87.69 | 91.39 | 99.00 | 99.30 | 99.30 |
| | STAN-Dev | 80.04 | 84.49 | 90.02 | 98.72 | 99.51 | 99.60 |
| | STAN-Small | 80.05 | 85.11 | 88.90 | 97.11 | 97.38 | 97.38 |
| | STAN-Large | 72.17 | 75.74 | 79.25 | 85.38 | 85.82 | 85.86 |
| | HashSet-Manual | 56.71 | 68.54 | 78.22 | 91.53 | 94.00 | 94.37 |

Table 2: Baseline Model Performance on the datasets. Accuracies improve as n reaches 10. Hashformer is consistently performing better than MPNR, but the performance of both the models is poorer on HashSet dataset as compared to other datasets.

| Architecture | % containing named entities | % containing non-English tokens |
|---|---|---|
| MPNR | 77.17 | 17.61 |
| Hashformer | 77.57 | 33.64 |

Table 3: Analysis of incorrectly segmented hashtags in HashSet - Manual for n=10. A majority of the incorrectly segmented hashtags contain named entities.

For each dataset, we generate top-10 segmentations. We evaluate the models using accuracy measure. A sample is classified as correct if the correct segmentation figures in top-$n$ segments, where $n$ ranges from 1 to 10. Table 2 shows the results for the aforementioned datasets and models.

Both, MPNR and Hashformer, perform well for BOUN, $STAN_{dev}$, $STAN_{small}$, $STAN_{large}$. Hashformer consistently outperforms MPNR across datasets. Accuracies improve as $n$ approaches 10.

On the HashSet-Manual dataset, however, performance of both models degrades substantially. Degradation in MPNR is much starker compared to Hashformer. We conjecture that this is due to the fact that MPNR relies on statistical LMs, which have lower coverage compared to the transformer-based LMs used by Hashformers.

From a utility perspective, segmentation is useful in downstream task only if the model gives higher accuracy for n=1, i.e., if the first segmentation is the correct one. We carry out error analysis for n=1. A reason for the poor performance of SOTA models on the HashSet

[2] https://github.com/ruanchaves/hashformers
[3] https://github.com/mounicam/hashtag_master

- Manual dataset is the presence of named entities and non-English tokens in the hashtags (see Table 3).

For $STAN_{dev}$, $STAN_{small}$, $STAN_{large}$ and BOUN, information about named entities and non-English tokens is not present, but manual error analysis on these datasets shows that, for MPNR, incorrectly segmented hashtags contain named entities. Examples of such hashtags are #GoViks, #10ThingsImAttractedToNiall, etc. We also noticed that hashtags containing numerals and abbreviations were also missegmented consistently across datasets. Examples of incorrectly segmented hashtags that contain numerals are #Scholar360, #Pasikatan2013, mirzapur2, etc. Many hashtags contain abbreviations like #cplt2013, #dream11iplfinal, #IHMFL, etc.

In datasets, apart from HashSet, few hashtags also contain underscores, which are a clear signal to segment. But MPNR and Hashformer utilize Language models to generate candidate segments, even for hashtags which have underscores in them. We argue that such hashtags should be handled automatically by splitting at underscores instead of relying on large models, which are an obvious over-kill. e.g., #What_A_Legend, #much_love, #weather_me, etc.

We noticed that few hashtags written in camel case were segmented incorrectly. On average, 17.8% of total hashtags written in camel case were incorrectly

**HashSet - Distant - Sampled - Lower cased**

| | n = 1 | n = 2 |
|---|---|---|
| MPNR | 45.04 | 50.45 |
| Hashformer | 47.43 | 58.59 |

**HashSet - Distant - Sampled**

| | n = 1 | n = 2 |
|---|---|---|
| Maddela et al. | 50.06 | 52.12 |
| Hashformer Rodriguez et al | 72.47 | 77.12 |

Table 4: Baseline Model Performance on HashSet - Distant dataset. The performance of the camel cased dataset is better than the lower cased dataset.

segmented by MPNR for n=1, and 15.8% were mis-segmented by Hashformer for n=1. The lower error rate in camel cased hashtags is indicative of the fact that camel case points in a hashtag are a strong signal for segmentation and are relatively easy for model to classify. Automatic splitting of hashtags at camel case points before feeding it a model is advantageous. For robust estimation of segmentation models, we removed any such camel cased hashtag from the HashSet-Manual dataset.

Since the camel cased hashtags are a strong signal for segmentation, we sample 20,000 camel cased hashtags from HashSet - Distant dataset. We kept both the camel cased version and the lower cased hashtag in the dataset. In Table 4 we compare accuracy for lower cased vs. camel cased hashtags, and both models gain substantially from camel case information, Hashformer model more so. This observation validates our hypothesis that camel cased hashtags are relatively easy for models to segment.

## 5. Discussion

We present the HashSet dataset, using both manually annotated and automatically generated loosely supervised data using hashtag patterns. We showed that when the hashtags are sourced from different collections of data, the performance of current SOTA models drops. To analyze the source of errors, we use the named entity annotations and non-English token annotations and show that the errors are predominantly in hashtags that have named entities, abbreviations, and numerals. A named entity recognizer that works on an unsegmented level could be useful, and we leave that as part of our future work.

Hashtags from different geographical locations will reflect named entities, different language preferences. We argue that datasets used to benchmark hashtag segmentation algorithms should reflect the same. The hashtags collection is sourced from Indian cities and collection of Indian election, hence named entities are from Indian origin, and the non-English tokens belong to Indian languages, with the majority being romanized Hindi tokens.

For the HashSet-Distant dataset, the patterns used to segment hashtags have high coverage, but there are certain edge cases where the hashtag might be incorrectly segmented. The erroneous cases that we noticed were caused by misspellings. However, the utility of splitting hashtags at camel case points before feeding to the model would nevertheless be useful and help in getting correct segments.

## 6. Bibliographical References

## 7. Language Resource References